\documentclass{llncs}

\usepackage[pdftex]{xcolor}
\usepackage[pdftex]{hyperref}
\usepackage{listings}
\usepackage{xspace}

\newcommand{\protege}{Prot\'eg\'e\xspace}
\newcommand{\webprotege}{Web-\protege}

\lstMakeShortInline[]|

\begin{document}

\title{The Semantic Web takes Wing: Programming Ontologies with Tawny-OWL}
\author{Phillip Lord}
\institute{School of Computing Science, Newcastle University}
\maketitle

\begin{abstract}
  The Tawny-OWL library provides a fully-programmatic environment for ontology
  building; it enables the use of a rich set of tools for ontology
  development, by recasting development as a form of programming. It is built
  in Clojure -- a modern Lisp dialect, and is backed by the OWL API. Used
  simply, it has a similar syntax to OWL Manchester syntax, but it provides
  arbitrary extensibility and abstraction. It builds on existing facilities
  for Clojure, which provides a rich and modern programming tool chain, for
  versioning, distributed development, build, testing and continuous
  integration. In this paper, we describe the library, this environment and
  the its potential implications for the ontology development process.
\end{abstract}

\section{Introduction}
\label{sec:introduction}

Ontology building remains a difficult and demanding task. Partly this is
intrinsic, but partly stems from the tooling. For example, while ontology
editors like \protege\cite{greycite2912} do allow manual ontology
development, they are not ideal for automation or template-driven development;
for these reasons languages such as
OPPL\cite{aranguren_Stevens_Antezana_2009} have been developed; these allow a
slightly higher-level of abstraction over the base OWL axiomatisation. However,
they involve a move away from OWL syntax, which in turn requires integration
into which ever environment the developers are using. There has also been
significant interest in collaborative development of ontologies, either using
collaborative development tools such as
Web-Protege\cite{tudorache_webprotege:2011}, or through copy-modify-merge
versioning\cite{iz_Grau_Horrocks_Berlanga_2011}.

In this work, we\footnote{Plurals are used throughout, and do not indicate
  multiple authorship.} take an alternative approach. Instead of developing
tools for ontology development, many of which are similar or follow on from
software development tools, we attempt to recast ontology development as a
software engineering problem, and then just use the standard tools that exist
for software engineering. We have achieved this through development of a
library, named \textit{Tawny OWL}, that at its simplest operates as a domain
specific language for OWL, while still retaining the full capabilities of a
modern programming language with all this entails. We demonstrate the
application of this library to a standard exemplar - namely the Pizza
Ontology\cite{greycite3196}, as well as several other scenarios. Finally, we
consider the implications of this approach for enabling collaborative and more
agile forms of ontology development.

\section{Requirements}
\label{sec:tool}

Interaction between OWL and a programming API is not a new idea. For example,
OWL2Perl\cite{Kawas_Wilkinson_2010} allows generation of Perl classes from an
OWL Ontology, while the OWL API allows OWL ontology development in
Java\cite{bechhofer03:_cookin_owl_api}. The OWL API, however, is rather
unwieldy for direct ontology development; for example, it has a complex type
hierarchy, indirect instantiation of objects through factories, and a set of
change objects following a command design pattern; while these support one of
its original intended use case -- building a GUI -- they would make direct
ontology development cumbersome. One response to this is
Brain\cite{croset2013,greycite8021}, which is a much lighter weight facade
over the OWL API also implemented in Java. Brain is, effectively, type-less as
expressions are generated using Strings; the API distinguishes between OWL
class creation (\texttt{addClass}) and retrieval (\texttt{getClass}), throwing
exceptions to indicate an illegal state. While Brain is useful, it is not
clear how an ontology should be structured in Java's object paradigm, and it
suffers the major drawback of Java -- an elongated compile-test-debug cycle,
something likely to be problematic for interactive development as the ontology
increases in size. 

For programmatic ontology development, we wanted a much more interactive and
dynamic environment; something equivalent to the R environment for statistics,
where the ontology could be explored, extended and reworked on-the-fly,
without restarting. For this reason we choose to build in Clojure; this
language is a modern Lisp derivative with many attractive features: persistent
data structures; specialised mechanisms for storing state. It suffers somewhat
from being built on the Java Virtual Machine (JVM) --in particular this gives
it a rather slow start-up time -- however, in this case, it was a key reason
for its use. Interoperability with the JVM is integrated deeply into Clojure
which makes building on top of the OWL API both possible and convenient. Like
all lisps, Clojure has three other advantages: first, it is untyped which, in
common with Brain, in this context, we consider to be an advantage\footnote{We
  do not argue that type systems are bad; just that the are less appropriate
  in this environment}; second, it is highly dynamic -- almost any aspect of
the language can be redefined at any time -- and it has a full featured
read-eval-print-loop (REPL); finally, it has very little syntax, so libraries
can manipulate the look of the language very easily. Consider, for example, a
simple class definition as shown in Listing\ref{lst:class}, taken from a pizza
ontology available at \url{https://github.com/phillord/tawny-pizza}. The
syntax has been designed after Manchester syntax\cite{greycite2216}.

\begin{lstlisting}[caption={A basic class definition},label=lst:class]
(defclass Pizza
   :label "Pizza"
   :comment 
"An over-baked flat bread with toppings, originating from Italy."
)
\end{lstlisting}

A more complex definition shows the generation of restrictions and anonymous
classes. 

\begin{lstlisting}[caption={A Cheesy Pizza},label=lst:cheesy]
(defclass CheesyPizza
  :equivalent
  (owland Pizza
           (owlsome hasTopping CheeseTopping)))
\end{lstlisting}

These definitions bind a new symbol (|Pizza| and |CheesyPizza|) to a OWLAPI
Java object. These symbols resolve as a normal Var does in Clojure. Strictly,
this binding is not necessary (and can be avoided if the user wishes),
however this provides the same semantics as Brain's |addClass| and |getClass|
-- classes, properties, etc must be created before use; this is a valuable
feature protecting against typing errors\cite{greycite3001}.

\subsection{Lisp Terminology}
\label{sec:lisp-terminology}

\newcommand{\eg}{\textit{e.g.}\xspace}

Here we give a brief introduction to Clojure and its terminology. Like all
lisps, it has a regular syntax consisting of parenthesis delimited |(lists)|,
defining an \textit{expression}. The first element is usually a
\textit{function}, giving lisps a prefix notation. Elements can be literals,
such as strings \eg |"Pizza"|, \textit{symbols} \eg |defclass| or
\textit{keywords} \eg |:equivalent|. Symbols resolve to their values, keywords
resolve to themselves, and literals are, well, literal. Unlike many languages,
these constructs are directly manipulable in the language itself which
combined with \textit{macros} enable extension of the language.

\section{A Rich Development Environment}
\label{sec:devel-envir}

There are a dizzying array of ontology development tools
available\cite{greycite8023}. Probably the most popular is \protege; while it
provides a very rich environment for viewing and interacting with an ontology,
it lacks many things that are present in most IDEs. For instance, it lacks
support for version control or adding to ChangeLogs; it is not possible to
edit documentation along side the ontology; nor to edit code in other
languages when, for instance, driving a build process, or using an ontology in
an application.

We have previously attempted to work around this problem by providing support
for Manchester syntax -- \textit{OMN} -- within Emacs through
\textit{omn-mode}\cite{greycite2218}; while this provides a richer
general-purpose environment, the ontology environment is comparatively poor.
In particular, only syntactic completion is available, there is no support for
documentation look-up, beyond file navigation. Finally, we used \protege (and
the OWL API) to check syntax, which required a complete re-parse of the file,
and with relatively poor feedback from \protege when errors
occurred\footnote{This is not a criticism of the \protege interface; it was
  not designed to operate on hand-edited files}.

With tawny, using a general purpose programming language, a richer development
environment comes nearly for free. In this paper, we describe the use within
Emacs; however, support for Clojure is also available within Eclipse,
IntelliJ, Netbeans and other environments\cite{greycite8024}. Compared with
direct editing of OMN files, this provides immediate advantages. The use of
paren delimiters makes indentation straight-forward, well-defined, and
well-supported; advanced tools like paredit ensures that expressions are
always balanced. Clojure provides a REPL, and interaction within this allows
more semantic completion of symbols even when they are not syntactically
present in the buffer\footnote{We follow Emacs terminology here -- a
  \textit{buffer} is a file being edited}, which is common when using levels
of abstraction (Section~\ref{sec:high-levels-abstr}) or external OWL files
(Section~\ref{sec:owl-files-semantics}). Syntax checking is easy, and can be
performed on buffer, marked region or specific expression. New entities can be
added or removed from the ontology on-the-fly without reloading the entire
ontology, enabling progressive development. We have also provided support for
documentation look-up of OWL entities; this is hooked into Clojure's native
documentation facility, so should function within all development
environments. We do not currently provide a rich environment for browsing
ontologies, except at the code level; however, \protege works well here,
reloading OWL files when they are changed underneath it. Similarly, omn-mode
can be used to view individual generated OMN files.

\section{Supporting Higher Levels of Abstraction}
\label{sec:high-levels-abstr}

Most ontologies include a certain amount of ``boilerplate'' code, where many
classes follow a similar pattern. Tools such as OPPL were built specifically
to address this issue; with tawny, the use of a full programming language,
makes the use of levels of abstraction above that in OWL straight-forward. We
have used this in many areas of Tawny; at its simplest, by providing
convenience macros. For example, it is common-place to define many subclasses
for a single superclass; using OMN each subclass must describe its superclass.
Within tawny, a dynamically-scoped block can be used as shown in
Listing~\ref{lst:subclass}. As shown here, disjoint axioms can also be
added\cite{greycite3875}; and, not used here, covering
axioms\cite{greycite1296}. The equivalent OMN generated by these expressions
is also shown in Listing~\ref{lst:subclassomn}.

\begin{lstlisting}[caption={Subclass Specification},label=lst:subclass]
(as-disjoint-subclasses
 PizzaBase
 
 (defclass ThinAndCrispyBase
   :annotation (label "BaseFinaEQuebradica" "pt"))

 (defclass DeepPanBase
   :annotation (label  "BaseEspessa" "pt")))
\end{lstlisting}

\begin{lstlisting}[caption={Subclasses in OMN},label=lst:subclassomn]
Class: piz:ThinAndCrispyBase
    Annotations: 
        rdfs:label "BaseFinaEQuebradica"@pt
    SubClassOf: 
        piz:PizzaBase
    DisjointWith: 
        piz:DeepPanBase

Class: piz:DeepPanBase
    Annotations: 
        rdfs:label "BaseEspessa"@pt,
    SubClassOf: 
        piz:PizzaBase
    DisjointWith: 
        piz:ThinAndCrispyBase
\end{lstlisting}

It is also possible to add suffixes or prefixes to all classes created within
a lexical scope. For example, we can create classes ending in \lstinline|Topping| as
shown in Listing~\ref{lst:suffix}. While similar functionality could be
provided with a GUI, this has the significant advantage that the developers
intent remains present in the source; so subsequent addition of new toppings
are more likely to preserve the naming scheme. 

\begin{lstlisting}[caption={Adding Suffixes},label=lst:suffix]
(with-suffix Topping
   (defclass GoatsCheese)
   (defclass Gorgonzola)
   (defclass Mozzarella)
   (defclass Parmesan))
\end{lstlisting}

Tawny also includes initial support for ontology design patterns; in
particular, we have added explicit support for the value
partition\cite{rector2005}. This generates classes, disjoints and properties
necessary to fulfil a pattern, but is represented in Tawny succinctly
(Listing~\ref{lst:valuep})

\begin{lstlisting}[caption={A Value Partition},label=lst:valuep]
(p/value-partition
 Spiciness
 [Mild
  Medium
  Hot])
\end{lstlisting}

While some abstractions are generally useful, an important advantage of a
full-programmatic language for OWL is that abstractions can be added to any
ontology including those which likely to be useful only within a single
ontology. These can defined as functions or macros in the same file as their
use. For example, within the pizza ontology, Listing~\ref{lst:named} generates
two pizzas -- in each case the pizza class comes first, followed by
constituent parts; a closure axiom is added to each pizza. As well, as being
somewhat more concise than the equivalent OMN, this approach also has the
significant advantage that it is possible to change the axiomatisation for all
the named pizzas by altering a single function; this is likely to increase the
consistency and maintainability of ontologies.

\begin{lstlisting}[caption={Generating Named Pizzas},label=lst:named]
(generate-named-pizza
 [MargheritaPizza MozzarellaTopping TomatoTopping]

 [CajunPizza MozzarellaTopping OnionTopping PeperonataTopping
  PrawnsTopping TobascoPepperSauce TomatoTopping]
\end{lstlisting}

\section{Separating Concerns for Different Developer Groups}
\label{sec:separation-concerns}

One common requirement in ontology development is a separation of concerns;
different contributors to the ontology may need different editing
environments, as for instance with RightField or
Populous\cite{Jupp_Wolstencroft_Stevens_2011}. Tawny enables this approach
also; here, we describe how this enables internationalisation. Originally, the
pizza ontology had identifiers in English and Portuguese but, ironically, not
Italian. While it would be possible to have a translator operate directly on a
tawny source file, this is not ideal as they would need to need to embed their
translations within OWL entity definitions as shown in
Listing~\ref{lst:subclass}; this is likely to be particularly troublesome if
machine assisted translation is required due to the non-standard format. We
have, therefore added support with the \textit{polyglot} library. Labels are
stored in a Java properties file (Listing~\ref{lst:italianres}) and are loaded
using a single Lisp form (Listing~\ref{lst:italian}). Tawny will generate a
skeleton resources file, with no translations, on demand, and reports missing
labels to the REPL on loading.

\begin{lstlisting}[caption={Italian Resources},label=lst:italianres]
AnchoviesTopping=Acciughe Ingredienti
ArtichokeTopping=Carciofi Ingredienti
AsparagusTopping=Asparagi Ingredienti 
\end{lstlisting}

\begin{lstlisting}[caption={Loading Multi-Lingual Labels},label=lst:italian]
(tawny.polyglot/polyglot-load-label 
  "pizza/pizzalabel_it.properties" "it")
\end{lstlisting}

Currently, only loading labels is supported in this way, but extending this to
comments or other forms of annotation is possible. While, in this case, we are
loading extra-logical aspects of the ontology from file, it would also be
possible to load logical axioms; for instance, named pizzas
(Section~\ref{sec:high-levels-abstr}) could be loaded from text file,
spreadsheet or database.

\section{Collaborative and Distributed Development}
\label{sec:vers-coll}

Collaborative development is not a new problem; many software engineering
projects involve many developers, geographically separated, in different time
zones, with teams changing over time. Tools for enabling this form of
collaboration are well developed and well supported. Some of these tools are
also available for ontology development; for instance, \webprotege enables
online collaborative editing. However, use of this tool requires installation
of a bespoke Tomcat based server, nor does it yet support offline, concurrent
modification\cite{tudorache_webprotege:2011}.

Alternatively, the ContentCVS system does support offline concurrent
modification. It uses the notion of structural equivalence for comparison and
resolution of conflicts\cite{iz_Grau_Horrocks_Berlanga_2011}; the authors
argue that an ontology is a set of axioms. However, as the named suggests,
their versioning system mirrors the capabilities of CVS -- a client-server
based system, which is now considered archaic.

For tawny, the notion of structural equivalence is not appropriate; critically,
it assumes that an ontology is a \emph{set} of axioms. This is not true with
tawny, for two reasons: first, tawny requires definition of classes before
use, so source code cannot be arbitrarily re-ordered; secondly, even where
this is not the case, only the ontology axioms are a set. Programmer intent is
often represented through non-axiomised sections of the code -- whitespace,
indentation and even comments which may drive a ``literate'' development
approach. A definition of a difference based purely on axiomatisation cannot
account for these differences; the use of a line-oriented syntactic diff will.

We argue here that by provision of an attractive and well-supported syntax, we
do not need to provide specific collaborative tooling. Tawny itself has been
built using distributed versioning systems (first mercurial and then git).
These are already advanced systems supporting multiple workflows including
tiered development with authorisation, branching, cherry-picking and so on. 
While ontology-specific tooling have some advantages, it is unlikely to
replicate the functionality offered by these systems, aside from issues of
developer familiarity.

Later, we also describe support for testing, which can also ease the
difficulty of collaborative working (Section~\ref{sec:testing-ci}).

\section{Enabling Modularity}
\label{sec:naming-spac-distr}

Tawny provides explicit support for name spacing and does this by building on
Clojure's namespace support. It is possible to build a set of ontologies
spread across a number of different files. Normally, each file contains a
single namespace; tawny mirrors this, with each namespace containing a single
ontology, with a defined IRI.

OWL itself does not provide a distribution mechanism for ontologies; the IRI
of an ontology does not need to resolve. In practice, this is often a
distribution mechanism; by default \protege will check for resolution if other
mechanisms fail; OBO ontologies, for example, are all delivered from their
IRI.
 
In contrast, Tawny builds on the Clojure environment; most projects are built
using the Leiningen tool which, in turn, uses the repository and dependency
management from Maven. When building the Pizza ontology in Tawny, the build tool
will fetch Tawny itself, the OWL API and HermiT, and their dependencies.
Ontological dependencies can be fetched likewise. Maven builds come with a
defined semantics for versioning, including release and snapshot
differentiation. A key advantage of this system is that multiple versions of a
single ontology can be supported, with different dependency graphs.

\section{Coping With Semantics Free Identifiers}
\label{sec:owl-files-semantics}

Importing one ontology from another is straight-forward in tawny. However, not
all ontologies are developed using tawny; we need to be able interact with
external ontologies only accessible through an OWL file. Tawny provides
facilities for this use-case: the library reads the OWL file, creates symbols
for all entities\footnote{It is possible to choose a subset}, then associates
the relevant Java object with this symbol. This approach is reasonably
scalable; tawny can import the Gene Ontology within a minute on a desktop
machine. Clojure is a highly-dynamic language and allows this form of
programmatic creation of variables as a first-class part of the language; so
an ontology read in this way functions in every sense like a tawny native
ontology. Ontology classes can be queried for their documentation,
auto-completion works and so forth.

However, there is a significant problem with this import mechanism. Tawny
must create a symbol for each OWL entity in the source ontology. By default,
tawny uses the IRI fragment for this purpose; while Clojure symbols have a
restricted character set which is not the same as that of the IRI fragment, in
practice this works well. However, this is unusable for ontologies built
according to the OBO ontology standard, which uses semantics-free, numeric
identifiers such as |OBI_0000107|. While this is a valid Clojure symbol, it is
unreadable for a developer. This issue also causes significant difficulties
for ontology development in any syntax; OMN is relatively human-readable but
ceases to be so when all identifiers become numeric. We have previously
suggested a number of solutions to this problem either through the use of
comments or specialised denormalisations\cite{greycite1402}, or through the
addition of an |Alias| directive providing a mapping between numeric and
readable identifier\cite{greycite1327}. However, all of these require changes
to the specification and tooling updates, potentially in several syntaxes.

For tawny, we have worked around this problem by enabling an arbitrary mapping
between the OWL entity and symbol name \cite{greycite4428}. For OBO
ontologies, a syntactic transformation of the |rdfs:label| works well. Thus,
|OBI_0000107| can be referred to as |provides_service_consumer_with|, while
|BFO_0000051| becomes the rather more prosaic |has_part|. 

While this solves the usability problem, it causes another issue for ontology
evolution; the label is open to change, independently of any changes in
semantics; unfortunately, any dependent ontology built with tawny will break,
as the relevant symbol will no longer exist. This problem does not exist for
GUI editors such as \protege because, ironically, they are not WYSIWYG -- the
ontology stores an IRI, while the user sees the label; changes to labels
percolate when reloading the dependent ontology. Tawny provides a solution to
this; it is possible to \textit{memorise} mappings between symbols and IRIs at
one point in time\cite{greycite7845}. If the dependency changes its label,
while keeping the same IRI, Tawny will recognise this situation, and generate
a \textit{deprecated} symbol; dependent ontologies will still work, but will
signal warnings stating that a label has changed and suggesting appropriate
updates. Currently these must be performed manually, although this could be
automated.

\section{Enabling Unit Testing and Continuous Integration}
\label{sec:testing-ci}

Unit testing is a key additions to the software development process which has
enabled more agile development. Adapting this process for ontology development
has previously been suggested\cite{Vr06unittests}, and implemented as a plugin
to \protege\cite{greycite8019}. To add this capability to tawny, we have
integrated reasoning; at the time of writing, only
ELK\cite{kazakov12:_elk_reason} is available as a maven resource in the Maven
Central repository, therefore we have developed a secondary maven build for
HermiT which allows use of this reasoner
also\cite{greycite4427}\footnote{Available at
  http://homepages.cs.ncl.ac.uk/phillip.lord/maven/, or on Github}, so both
these reasoners are available for use; others can be added trivially as they
are \textit{mavenised}. A number of test frameworks exist in Clojure; here we
use |clojure.test|. As shown in Listing~\ref{lst:unit}, we check that various
inferences have occurred (or not as appropriate), using the |isuperclass?|
predicate. We have also added support for ``probe'' classes. In our second
test, we use the |with-probe-entities| macro; this adds a subclass of
|VegetarianPizza| and |CajunPizza| -- as the latter contains meat, this should
result in an incoherent ontology if both definitions are correct; probe
entities are automatically removed by the macro, returning the ontology to its
previous state, ensuring independence of tests.

\begin{lstlisting}[caption={Unit Testing a Pizza Ontology},label=lst:unit]
(deftest CheesyShort
  (is (r/isuperclass? p/FourCheesePizza p/CheesyPizza))
  (is (r/isuperclass? p/MargheritaPizza p/CheesyPizza))
  (is 
   (not (r/isuperclass? p/MargheritaPizza p/FourCheesePizza))))

(deftest VegetarianPizza
  (is 
   (r/isuperclass? p/MargheritaPizza p/VegetarianPizza))

  (is 
   (not
    (o/with-probe-entities
      [c (o/owlclass "probe"
                     :subclass p/VegetarianPizza p/CajunPizza)]
      (r/coherent?)))))
\end{lstlisting}

The use of Unit testing in this way has implications beyond simple ontology
development; it also allows a richer form of continuous integration where
dependent ontologies can be developed by independent developers, but
continuously checked for breaking changes. The tawny pizza ontology, for
example, is currently being tested using
Travis\footnote{\url{http://travis-ci.org}}. Unlike, other ontology CI
systems\cite{greycite2899}, this requires no installation, integrates directly
with the DVCS in use. It is also useful for integration with software that
operates on the ontology; for example, both our version of Hermit, the OWL API
and tawny-owl are built and tested using this tool.

\section{Discussion}
\label{sec:discussion}

In this paper, we have described Tawny, a library which enables the user to
develop ontologies, using the tooling and environments that have long been
available to programmers. Although they both involve producing artifacts with
strong computational properties ontology development and software engineering
have long been disjoint. This has significant negative impacts; there are far
more programmers than knowledge engineers, and as a result the tooling that
they use is far better developed. Tawny seeks to address this, not by
providing richer tools for ontology development, but by recasting ontology
development as a form of programming. 

By allowing knowledge engineers to use any level of abstraction that they
choose, tawny can also improve current knowledge engineering process
significantly. It can help to remove duplication, for example, in class names.
It can clearly delineate disjoint classes protecting against future additions;
this helps to address a common ontological error\cite{rector_owl_2004}. It is
also possible to model directly using common ontology design patterns
generating many axioms in a succinct syntax. Bespoke templates can be built
for a specific ontology; this mirrors functionality of tools like
OPPL\cite{aranguren_Stevens_Antezana_2009}, but uses the power of a full
programming language and environment. Trivially, for example, tawny can log
its activity and comes with debugger support.

Of course, direct use of a programmatic library like tawny is not suitable for
all users; however, even for these users a library like tawny could be useful.
It is straight-forward to integrate ontologies developed directly with tawny
as a DSL with knowledge stored in other formalisms or locations. In this
paper, we described loading multi-lingual labels from properties files,
isolating the translator from the ontology, and interacting with OWL files
generated by another tool. It would also be possible to load axioms from a
database or spreadsheet, using existing JVM libraries.

While with tawny, we have provided a programmatic alternative to many
facilities that exist in other tools, we also seek to provide tooling for a
more agile and reactive form of ontology development. Current waterfall
methodologies, exemplified by BFO-style realism lack agility, failing to
meet the requirement for regular releases to address short-comings, as has
been seen with both BFO 1.1\cite{greycite8036} and BFO 2.0\cite{greycite7916}. 
Likewise, the OBO foundry places great emphasis on a review process which is,
unfortunately, backlogged\cite{obo_foundry_reorg} -- in short, as with
waterfall software methodologies, the centralised aspects of this development
model appear to scale poorly. 

Tawny uses many ready-made and well tested software engineering facilities:
amenability to modern DVCS, a versioning and release semantics, a test
framework and continuous integration. The provision of a test environment is
particularly important; while ontology developers may benefit from testing
their own ontologies, the ability to contribute tests to their ontological
dependencies is even more valuable. They can provide upstream developers
precise and executable descriptions of the facilities which they depend on;
this gives upstream developers more confidence that their changes will not have
unexpected consequences. When this does happen, downstream developers can
track against older versions of their dependencies, obviating the need for
co-ordination of updates; when they do decide to update, the re-factoring
necessary to cope with external changes will be supported by their own test
sets; finally, continuous integration will provide early warning if their own
changes impact others. In short, tawny provides the tools for a more pragmatic
and agile form of ontology development which is more suited to fulfilling the
changing and varied requirements found in the real
world\cite{reality_in_biology_2010}.

\bibliographystyle{splncs}
\bibliography{owled2013-tawny}

\end{document}